\documentclass{article}

\usepackage[preprint]{neurips_2026}

\usepackage[utf8]{inputenc} 
\usepackage[T1]{fontenc}    
\usepackage{hyperref}       
\usepackage{url}            
\usepackage{booktabs}       
\usepackage{amsfonts}       
\usepackage{nicefrac}       
\usepackage{microtype}      
\usepackage{xcolor}         
\usepackage{graphicx}
\usepackage{amsmath}
\usepackage{multirow}
\usepackage{wrapfig}
\usepackage{graphicx} 
\usepackage{xcolor}
\usepackage{algorithm}
\usepackage{algorithmic}
\usepackage{amsmath}
\usepackage{enumitem}

\usepackage{utfsym}
\usepackage[table]{xcolor}

\title{ActiveFly-Bench: Aligning Embodied Question Answering with Vision-Language-Action for Aerial Embodied Perception}

\author{%
  Weichen Zhang$^{1,*}$,
  Shiquan Yu$^{1,*}$,
  Yinan Zhu$^{1,*}$,
  Peizhi Tang$^{2}$,
  Shilong Ji$^{1}$,
  \\
  \textbf{Zhiyuan Deng}$^{1}$,
  \textbf{Tianyi Lyu}$^{1}$,
  \textbf{Haoyang Wang}$^{1}$,
  \textbf{Xin Zeng}$^{1}$,
  \textbf{Chen Gao}$^{1,\dagger}$,
  \\
  \textbf{Yong Li}$^{1}$,
  \textbf{Xinlei Chen}$^{1,\dagger}$\\
  $^{1}$Tsinghua University,
  $^{2}$Manifold AI,\\
  $^*$Equal contribution
$^\dagger$Corresponding author\\
  \texttt{zhangwc23@mails.tsinghua.edu.cn},
  \texttt{chgao96@gmail.com},\\
  \texttt{chen.xinlei@sz.tsinghua.edu.cn},
  \texttt{liyong07@tsinghua.edu.cn}
}

\begin{document}

\maketitle

\begin{abstract}
Unmanned Aerial Vehicles (UAVs) have emerged as promising embodied agents for perception due to their free 3D mobility. A key requirement for embodied perception is the ability to navigate to task-relevant regions, acquire informative observations, and complete the task through scene understanding.
Prior work has largely focused on either language-guided UAV control in the physical world or high-level scene understanding in cyberspace, leaving a clear gap between the two. In this work, we introduce ActiveFly-Bench, the first benchmark that aligns cyberspace and the physical world for UAV embodied perception. It decomposes the problem into three hierarchical tasks: Aerial Embodied Question Answering (Air-EQA) for cyberspace reasoning, Fine-grained Language-Guided UAV Control (FLUC) for physical-world interaction, and Observation Behavior Planning (OBP) to bridge the two domains. The datasets are collected from diverse simulated and real-world environments, supporting imitation learning for aerial vision-language-action tasks.
We further develop ActiveFly, a closed-loop UAV agent that integrates visual-language reasoning with fine-grained control, and deploy it on a physical UAV platform. Experiments with representative agentic frameworks show that current UAV agents still struggle with behavior planning and viewpoint adjustment in embodied perception. These results establish ActiveFly-Bench as a new testbed for embodied aerial intelligence.
We release datasets and reproducible codes in the anonymous project page: \url{https://lvmolvmo.github.io/ActiveFly/}.

\end{abstract}

\section{Introduction}
\label{sec:intro}



Bridging cyberspace and the physical world is a core capability of embodied AI. Language-guided embodied perception~\cite{liu2025aligning} is one of the representative tasks. Given a natural language instruction such as \textit{"Check what’s behind the wall"}, the agent must actively explore the environment, adjust its viewpoint, and answer the question based on the acquired visual information. Such requirements make UAVs with free 3D mobility~\cite{chen2024ddl,zhou2021fuel,zhou2020ego} a well-suited platform. 

Recent work has explored several related tasks in low-altitude aerial scenarios. A representative line of research is aerial vision-and-language navigation (VLN)~\cite{liu2023aerialvln,wang2024towards,gao2025openfly,lee2025citynav,lin2025openvln}, which originates from ground robotics~\cite{anderson2018vision,krantz2020beyond,chen2019touchdown,ku2020room}. These efforts primarily focus on improving the agent’s instruction-following capability, requiring UAVs to interpret natural language instructions to navigate to a target location or search for a target object. Another related direction is Embodied Question Answering (EQA)~\cite{das2018embodied,islam2024eqa,wijmans2019embodied}, in which an embodied agent answers open-vocabulary questions through active exploration of the environment.

While these tasks partially capture an agent’s ability to actively acquire information through interaction with the environment, they still have limitations in evaluating UAV intelligence in embodied perception.
(1) \textbf{Semantic gap between physical-world control and high-level cyberspace reasoning.} Current EQA benchmarks either omit the intermediate action execution process~\cite{zhao2025cityeqa,ren2024explore} or provide exploration trajectories~\cite{majumdar2024openeqa,jiang2025beyond}  that lack explicit task-driven behavioral motivation. On the other hand, aerial VLN~\cite{liu2023aerialvln,wang2025uav} focuses on action prediction but lacks a corresponding high-level EQA objective. Its long and structured instructions deviate from realistic human–UAV interaction, making it difficult to build EQA tasks on top of it. Therefore, existing benchmarks remain limited in bridging the physical world and cyberspace for embodied perception. (2) \textbf{Coarse-grained action control.} Existing language-guided UAV control tasks~\cite{chen2026aerialvla,xu2026aerialvla,wang2025uav} only require the UAV to reach a position near the target, while ignoring its final orientation. However, in practical perception tasks, a UAV needs not only to reach the target position but also to adjust its viewpoint to obtain a desired observation. (3) \textbf{Lack of real-world validation.} Most existing UAV language-guided interaction systems are evaluated in simulation. While some pioneering works~\cite{wu2025vla, wang2025uav} demonstrate real-world UAV control with short instructions, there is still a lack of real-world systems that support language-guided embodied perception.

\begin{figure}[t]
  \centering
  \includegraphics[width=\linewidth]{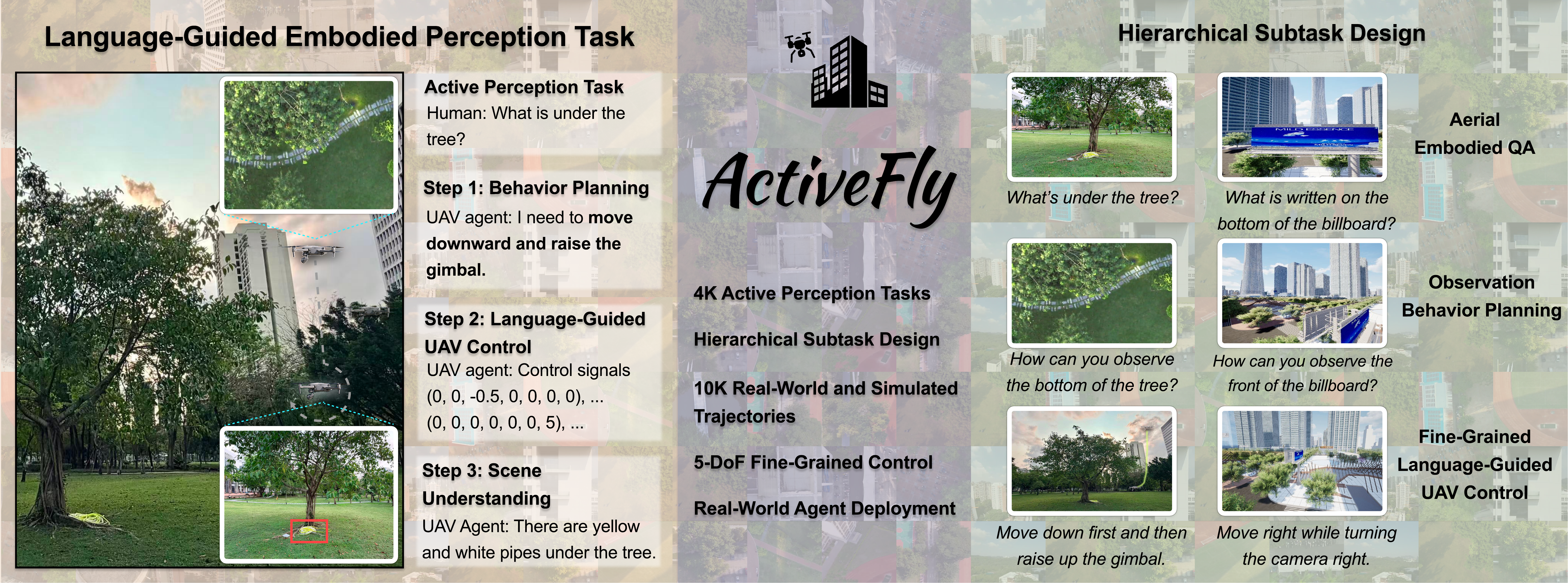}
  \vspace{-0.3cm}
  \caption{\textbf{Overview of ActiveFly-Bench.} The benchmark targets language-guided embodied perception for UAVs. It decomposes the task into three hierarchical and semantically aligned subtasks: Aerial Embodied Question Answering, Observation Behavior Planning, and Fine-Grained Language-Guided UAV Control. Given a high-level task such as “What is under the tree?”, the UAV first performs reasoning and planning to infer the observation behavior, then executes fine-grained control actions to reach an appropriate viewpoint, and finally answers the question based on the visual information.}
  \vspace{-0.3cm}
  \label{fig:cover}
\end{figure}

To address these limitations, we introduce \textbf{ActiveFly-Bench}, a benchmark designed to systematically evaluate UAV intelligence in language-guided embodied perception tasks.
We characterize such intelligence as the UAV’s ability to reason about appropriate observation behaviors from high-level instructions and current observations, generate executable control signals, and accomplish the task using visual information gathered during execution.
For example, given a task such as “check what’s under the tree,” the agent is expected to first infer the desired observation behavior (e.g., “descend and tilt the camera upward”), then generate a sequence of executable actions, and finally answer the question based on the visual observations.

Following this paradigm, ActiveFly-Bench comprises three hierarchical tasks for embodied perception: Aerial EQA (Air-EQA), Observation Behavior Planning (OBP), and Fine-grained Language-guided UAV Control (FLUC).
Air-EQA is a high-level task where the UAV agent cannot answer a question from its current aerial perspective, but adjust its viewpoint to acquire critical visual information.
OBP requires the agent to infer explicit observation behaviors conditioned on the question in Air-EQA.
FLUC further requires the agent to generate executable fine-grained actions based on the inferred observation behaviors. Compared with aerial VLN~\cite{liu2023aerialvln} and VLA~\cite{wang2025uav}, FLUC differs in two aspects: (1) It requires the UAV to reach an optimal viewpoint, constraining both final position and orientation. (2) It extends the action space to 5 DoF by jointly modeling UAV-body and camera-gimbal control, making it more realistic for real-world UAV platforms~\cite{dji_mavic_series}.
All UAV trajectories are collected from both real-world and simulated environments, covering three outdoor scenarios and one indoor scenario. Air-EQA, OBP, and FLUC are then constructed from the same trajectories, ensuring semantic alignment across the three tasks.


To validate whether a UAV agent can perform embodied perception tasks in real-world environments, we design the \textbf{ActiveFly} agent and deploy it on a physical UAV platform. We adopt a ground–drone collaborative framework similar to UAV-Flow~\cite{wang2025uav}, in which the UAV transmits its state and visual inputs to a ground station for inference and receives control feedback with low latency.

In summary, our contributions are as follows:

\begin{itemize}[leftmargin=*,partopsep=0pt,topsep=0pt]
\setlength{\itemsep}{0pt}
\setlength{\parsep}{0pt}
\setlength{\parskip}{0pt}
\item We introduce \textbf{ActiveFly-Bench}, a benchmark for evaluating UAV intelligence in language-guided embodied perception. It decomposes the task into hierarchical, semantically aligned subtasks, enabling the evaluation of reasoning, planning, and action execution in a unified framework.
\item We provide the first {fine-grained language-guided UAV control (FLUC)} dataset, which jointly models UAV-body and gimbal control for viewpoint-aware action execution. Built on FLUC, we further construct aerial embodied question answering (Air-EQA) and {Observation Behavior Planning (OBP)} to bridge high-level scene understanding and low-level control.
\item We develop the \textbf{ActiveFly} agent and systematically evaluate representative VLMs and VLAs on the benchmark. Through task-level and failure-mode analysis, we identify key bottlenecks in current UAV embodied perception systems and further validate the framework through real-world UAV deployment.
\end{itemize}

\begin{table}[t]
\centering
\small
\caption{\textbf{Comparison between existing benchmarks and ActiveFly-bench.}}
\setlength{\tabcolsep}{5pt}
\renewcommand{\arraystretch}{1.2}
\resizebox{\textwidth}{!}{%
\begin{tabular}{lcccccccccc}
\toprule
\multirow{2}{*}{Method}
& \multirow{2}{*}{Environment}
& \multirow{2}{*}{Data Source}
& \multicolumn{3}{c}{QA}
& \multicolumn{4}{c}{Language-guided action control}
& \multirow{2}{*}{Annotation} \\
\cmidrule(lr){4-6} \cmidrule(lr){7-10}
& & & Active & Planning & \# QA & QA-aligned instr. & Action space & Traj. feature & \# Traj & \\
\midrule
EQA-v1~\cite{das2018embodied} & Indoor & Simulation & \usym{2713} & \usym{2718} & 5.0k & \usym{2718} & 2 DoF & Short, fine-grained & - & Rule-based \\
MP3D-EQA~\cite{wijmans2019embodied} & Indoor & Real World & \usym{2713} & \usym{2718} & 1.1k & \usym{2718} & 2 DoF & Short, fine-grained & - & Rule-based \\
HM-EQA~\cite{ren2024explore} & Indoor & Real World & \usym{2713} & \usym{2718} & 500 & \usym{2718} & 2 DoF & - & - & VLM \\
OpenEQA~\cite{majumdar2024openeqa} & Indoor & Real World & \usym{2713} & \usym{2718} & 1.6k & \usym{2718} & 2 DoF & Short, fine-grained & 152 & Human \\
EXPRESS~\cite{jiang2025beyond} & Indoor & Real World & \usym{2713} & \usym{2718} & 2.0k & \usym{2718} & 2 DoF & Short, fine-grained & 777 & VLM\\
CityEQA~\cite{zhao2025cityeqa} & Outdoor & Simulation & \usym{2713} & \usym{2718} & 1.4k &  \usym{2718} & 4 DoF & - & - & Human \\
AerialVLN~\cite{liu2023aerialvln} & Outdoor & Simulation & - & - & - & \usym{2718} & 4 DoF & Long, coarse-grained & 100k & Human \\
OpenUAV~\cite{wang2024towards} & Outdoor & Simulation & - & - & - & \usym{2718} & 6 DoF & Long, coarse-grained & 12.6k & VLM \\
OpenFly~\cite{gao2025openfly} & Outdoor & Simulation & - & - & - & \usym{2718} & 4 DoF & Long, coarse-grained & 100k & VLM \\
UAV-Flow~\cite{wang2025uav} & Outdoor & Real. \& Sim. & - & - & - & \usym{2718} & 4 DoF & Short, fine-grained & 40k & Human  \\
\textbf{ActiveFly-Bench} & In. \& Outdoor & Real. \& Sim. & \usym{2713}  & \usym{2713}  & 1.3k & \usym{2713}  & 5 DoF & Short, fine-grained & 10k & Human  \\
\bottomrule
\end{tabular}%
}
\label{tab:benchmark_comparison}
\end{table}
 
\section{Related Work}
\subsection{Benchmark for EQA}
The EQA task~\cite{das2018embodied,majumdar2024openeqa,yu2019multi,li2026industryeqa,zhai2025multi,tan2023knowledge} typically requires agents to actively explore the environment by placing question-relevant semantics outside the initial field of view. However, many benchmarks~\cite{wijmans2019embodied,yu2019multi,cangea2019videonavqa} do not provide reference trajectories, making it difficult to disentangle exploration from reasoning based on QA accuracy alone. Others rely on pre-specified memory~\cite{majumdar2024openeqa,zhai2025memory,datta2022episodic}, which weakens their ability to reflect active perception. Although benchmarks such as EQA-v1~\cite{das2018embodied} and OpenEQA~\cite{majumdar2024openeqa} provide both QA pairs and reference trajectories, they still lack the intermediate reasoning process that links a question to a plausible exploration trajectory.
In addition, most existing EQA benchmarks are limited to closed indoor environments~\cite{tan2020multi,wu2024noisyeqa}, restricting their ability to evaluate embodied intelligence in open outdoor settings. CityEQA~\cite{zhao2025cityeqa} extends EQA to urban outdoor scenes, but its lack of reference trajectories and the non-uniqueness of valid solutions make exploration efficiency hard to assess. Finally, most EQA benchmarks remain confined to simulated environments, with little or no system-level validation in the real world.


\subsection{VLA for Aerial Navigation}
For UAVs, the primary objective of VLA tasks is to navigate to a target location based on language instructions. Early work formulates this problem as Aerial VLN~\cite{liu2023aerialvln,gao2025openfly,lee2025citynav,wang2025uav,xiao2025uav}, which mainly evaluates an aerial agent’s ability to follow long instructions and perform long-horizon, coarse-grained navigation. More recent efforts~\cite{liu2026indooruav,sun2026air}, such as VLA-AN~\cite{wu2025vla} and UAV-Flow~\cite{wang2025uav}, focus instead on short-horizon, fine-grained action control.
Although existing aerial VLA benchmarks have enabled deployment in the physical world and interaction with real environments, they still remain semantically disconnected from high-level tasks. A fundamental question is: after the agent follows an instruction and reaches a location, what can it actually do there?

To address this gap, we propose ActiveFly-Bench, which bridges high-level reasoning in cyber space and low-level execution in the physical world, enabling a more comprehensive evaluation of an agent’s active exploration and semantic reasoning capabilities in real-world environments.



\section{Problem Formulation}
\label{sec:prob_formu}
We decompose the language-guided embodied perception task into three fundamental and correlated subtasks: \textit{Observation Behavior Planning (OBP)}, \textit{Fine-grained Language-Guided UAV Control (FLUC)}, and \textit{Aerial Embodied Question Answering (Air-EQA)}.

Air-EQA is the final task to be completed by the UAV agent, and serves as the high-level objective that drives embodied perception. Specifically, an Air-EQA instance is defined as a tuple $(Q, A^*, O_0)$, where $Q$ is an open-vocabulary question, $A^*$ is the ground-truth answer, and $O_0$ denotes the agent's initial visual observation. The agent must explore the environment and produce an answer $A$ based on its historical observations $O_{0:N}$, which is formulated as
\begin{equation}
    A = \texttt{ActiveFlyAgent}(Q, O_{0:N}),
\end{equation}
where $N$ is the total exploration step.

OBP serves as an intermediate step for solving Air-EQA. It is defined as a tuple $(O_0, Q, L^*_{\text{ob}})$, where $L^*_{\text{ob}}$ denotes the observation behavior description annotated by humans. OBP evaluates whether the agent can translate high-level task intentions into task-relevant observation behaviors based on the current visual context. Formally, the agent predicts the observation behavior description as
\begin{equation}
    L_{\text{ob}} = \texttt{ActiveFlyAgent}(Q, O_0).
\end{equation}

FLUC corresponds to the low-level action execution stage conditioned on the observation behavior description from OBP. To better reflect real-world UAV embodied perception, unlike conventional UAV VLN~\cite{liu2023aerialvln} or VLA~\cite{wang2025uav} settings, we additionally introduce gimbal pitch control, enabling more fine-grained action execution and complex observation behaviors. The agent maps visual observations and textual instructions to embodied actions, similar to the standard VLA formulation~\cite{kim2024openvla,intelligence2025pi05,black2024pi0}:
\begin{equation}
    a_t = \texttt{ActiveFlyAgent}(L^{*}_{\text{ob}}, O_{t}, \mathbf{x}_t),
\end{equation}
where $a_t \in \mathbb{R}^5$ consisting of 3D translational motion, yaw, and pitch. After executing $N$ actions, the agent collects a history of observations $O_{0:N}$, which is used to answer the embodied question $Q$. 
\section{Dataset Collection and Validation}
\label{sec:data_gen}

\subsection{FLUC Data Collection}
\label{sec:fluc_gen}

\begin{figure}[t]
  \centering
  \includegraphics[width=\linewidth]{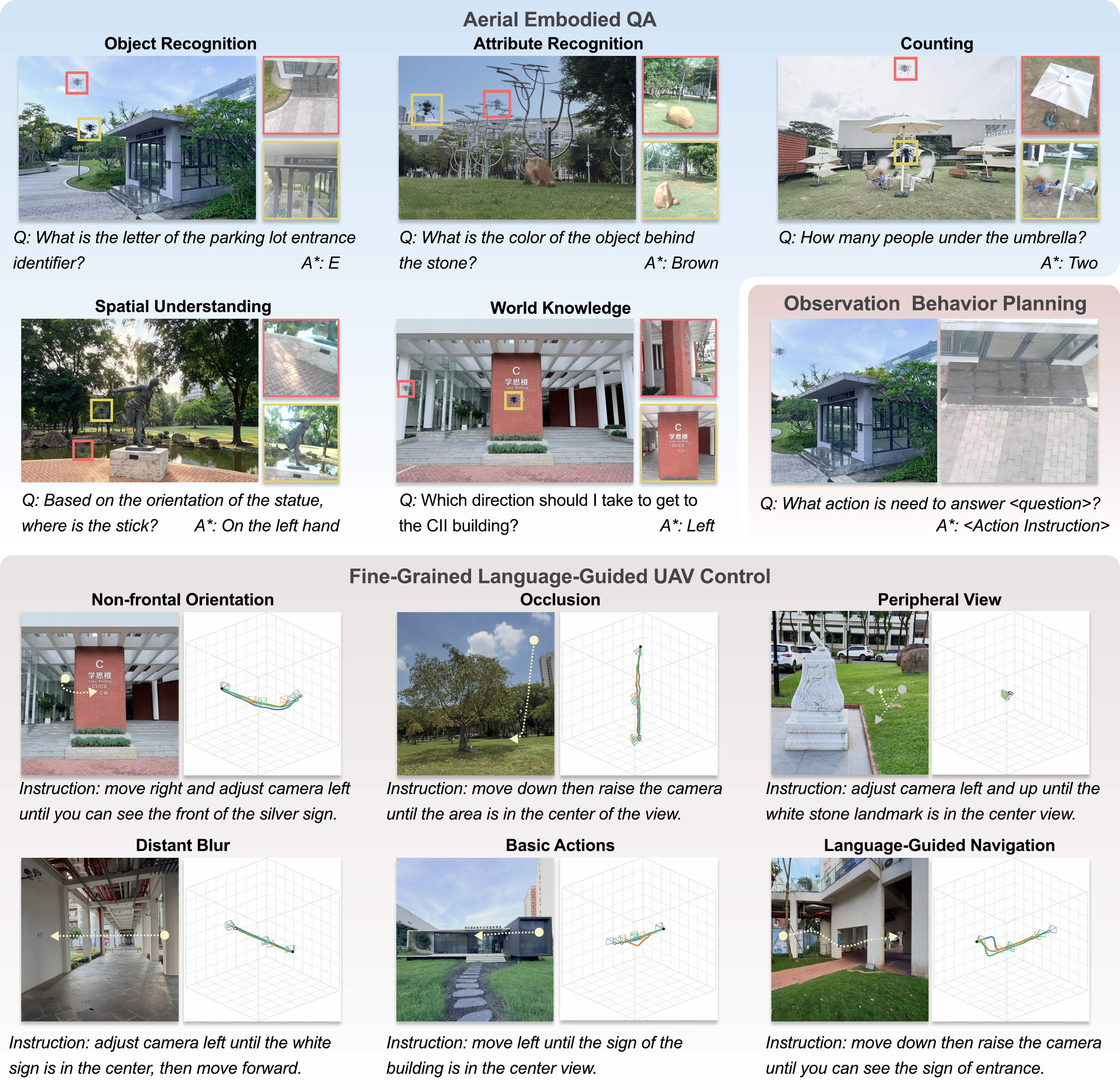}
  \vspace{-0.5cm}
  \caption{\textbf{Examples of the three tasks in ActiveFly-Bench.} Air-EQA spans five question categories: object recognition, attribute recognition, counting, spatial understanding, and world knowledge, with each example illustrated by the first and last frames. Observation Behavior Planning bridges Air-EQA and FLUC. FLUC includes 10 elementary and 4 complex active perception trajectories}
  \vspace{-0.5cm}
  \label{fig:task_illus}
\end{figure}


\textbf{High-Quality Trajectory Collection.}
Unlike conventional language-guided UAV navigation, which mainly requires reaching a target position via a short path, our task emphasizes human-like viewpoint adjustment under visually constrained conditions. To this end, each trajectory must satisfy three criteria: (1) it involves a clearly defined target object or region; (2) the target is not clearly observable from the initial position but becomes visible at the final position; and (3) the motion is consistent with human intuition. We therefore select semantically meaningful targets, such as billboards, landmarks, and open spaces, and define four representative observation-constrained scenarios: \emph{Occlusion}, \emph{Non-Frontal Orientation}, \emph{Peripheral View}, and \emph{Distant Blur}. For every trajectory, the initial observation must satisfy at least one constraint, while the final observation must provide a clear view of the target.

To collect high-quality trajectories, we employ experienced pilots to navigate UAVs to target viewpoints while jointly adjusting the UAV body and gimbal, so that the target can be observed with minimal movement. In addition to these four complex observation behaviors, we design ten elementary behaviors, including move forward/backward, move left/right, move up/down, turn left/right, and gimbal up/down, to support fine-grained 5-DoF control learning. By combining diverse targets with different constrained scenarios, we collect semantically diverse trajectories and first-person videos.
Finally, we collect 10k trajectories from one real-world campus scene, one real-world indoor scene, and two high-fidelity urban environments~\cite{gao2024embodiedcity,liu2023aerialvln}.

\begin{figure}[t]
  \centering
  \includegraphics[width=\linewidth]{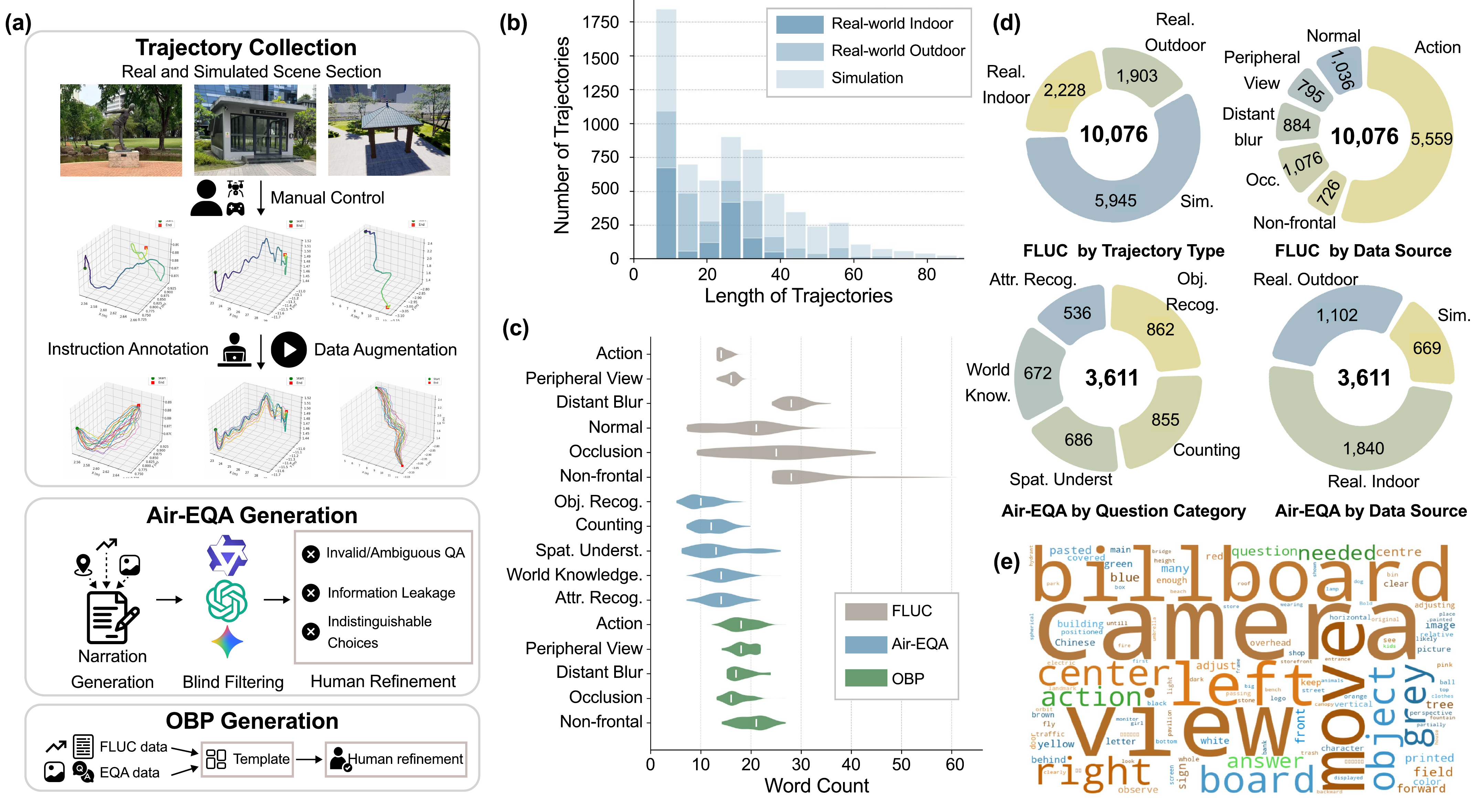}
  \vspace{-0.5cm}
  \caption{\textbf{Dataset Generation Pipeline and Statistics.} (a) The data generation pipeline of three tasks. (b) The distribution of FLUC trajectory lengths. (c) Word-count distributions of language annotations. (d) Dataset composition statistics. (e) A word cloud generated from the language annotations.} 
  \vspace{-0.8cm}
  \label{fig:data_stats}
\end{figure}

\textbf{Language Instruction Annotation} We adopt a hybrid annotation pipeline to generate language instructions for each trajectory and remove invalid samples. For elementary observation behaviors, we design instruction templates of the form "\texttt{<Action> until <target object> is in the center of the field of view}", which enables automatic generation of clear instructions by filling in the action type and target object.
In contrast, complex observation behaviors often involve both translational motion and orientation adjustment
For these cases, we employ experienced annotators to produce unambiguous instructions. 

\textbf{Data Augmentation} To support sample-efficient imitation learning, we draw inspiration from LIBERO~\cite{liu2023libero} and UAV-Flow~\cite{wang2025uav} and provide multiple demonstrations for each instruction based on existing trajectories. Specifically, we generate trajectories that exhibit similar observation behaviors but are not identical by perturbing the positions and orientations of waypoints along the original trajectory. Specifically, Gaussian noise is added to the start point, end point, and intermediate key waypoints to introduce slight deviations. A path planning algorithm is then applied to optimize a smooth trajectory, which is replayed by the UAV to automatically generate 5 to 40 demonstrations for each instruction.

\subsection{Air-EQA Data Collection}
\label{sec:eqa_gen}
\textbf{Question Generation.} We build Air-EQA samples from high-quality trajectories in the FLUC dataset. Given the first-person observations along each trajectory, human annotators generate question–answer pairs based on the changes in visual semantics.
We define five categories of Air-EQA questions to evaluate the UAV’s abilities in (1) object recognition (e.g., “What is under the tree?”), (2) attribute recognition (e.g., “What is the color of the object under the tree?”), (3) counting (e.g., “How many people are beneath the pavilion?”), (4) spatial reasoning (e.g., “What is on the left side of the red tree?”), and (5) world-knowledge reasoning, as illustrated in Figure~\ref{fig:task_illus}.
To ensure the embodied nature of the task, each question is constructed so that it cannot be answered from the initial observation alone, but becomes answerable from the target viewpoint.

\textbf{Data Validation.} Since some questions are prone to information leakage, their answers may be inferred directly from the question itself or the initial observation image. To address this issue, we adopt a blind filtering strategy by querying multiple commercial multimodal large language models, including GPT-5, Gemini, and Qwen, with the question and the initial observation. Samples correctly answered by all three models are discarded.
The dataset is then further refined by human reviewers, who filter out samples with ambiguous target references, open-ended answers, or questions that can be answered without UAV movement. We also refine distractor answer choices that cannot be ruled out solely from the context.

\subsection{Bridging the Gap Between FLUC and Air-EQA}
\label{sec:obp_gen}
Although Air-EQA requires the agent to actively explore the environment, it does not explicitly model how the exploration should be conducted, resulting in exploration behaviors that often lack human-like priors. To address this limitation, we construct the OBP dataset to guide the agent in reasoning about appropriate movements for completing Air-EQA tasks.

\textbf{OBP Data Generation} Given the FLUC dataset and the corresponding Air-EQA annotations, annotators construct OBP question–answer pairs by reformulating the Air-EQA questions and FLUC instructions. We first design a template of the form:
\texttt{Q: You are a UAV agent. You can control your motion, yaw, and gimbal pitch. How should you move to answer the question: <Air-EQA question>?}
\texttt{A: <FLUC instruction>}.
The generated OBP pairs are then refined by human annotators to remove information inconsistency and leakage. 

\subsection{Dataset Analysis}
\label{sec:data_analysis}


\textbf{Trajectory and Instruction Analysis.}
As depicted in Figure~\ref{fig:data_stats}, the proposed benchmark contains 10k FLUC trajectories, 1.3k Air-EQA pairs, and 1.3k OBP pairs. FLUC includes 6.2k simulated trajectories, 1.9k real-world indoor trajectories, and 1.9k real-world outdoor trajectories. Most trajectories range from 10 to 40 meters. The instruction lengths are mainly distributed between 15 and 30 words. Air-EQA questions are generally shorter, mostly ranging from 5 to 20 words. OBP is in one-to-one correspondence with Air-EQA,  with lengths mainly between 10 and 30 words.

\textbf{Dataset Splits.}
For fine-grained UAV control, the FLUC dataset is split into 80\% training data and 20\% testing data, with the same ratio maintained across all trajectory categories, yielding 8k training trajectories and 2k testing trajectories. In contrast, all Air-EQA and OBP pairs are reserved for evaluating scene understanding and planning capabilities.

\section{ActiveFly Agent}
\label{sec:agent}
In this section, we introduce \textbf{ActiveFly}, a closed-loop UAV system designed for language-guided active perception, and describe how its capabilities are comprehensively evaluated. We also present its real-world deployment on a physical UAV platform for active perception in real environments.

\textbf{Closed-loop System Design}
ActiveFly consists of a VLM~\cite{singh2025openai,comanici2025gemini,yang2025qwen3,liu2023visual} and a VLA~\cite{kim2024openvla,kim2025fine,intelligence2025pi05} module. As described in Section~\ref{sec:prob_formu}, given an embodied question, ActiveFly first uses the VLM to plan the observation behavior. Specifically, the initial observation and the question are provided as input to the VLM, which predicts a textual observation plan. This predicted plan is then used as the language instruction for the VLA model, together with the current observation and UAV state, to predict fine-grained UAV actions. After the agent executes the planned trajectory, it queries the VLM with $n$ images sampled from the observation history and the embodied question to obtain the final answer. In practice, we set $n=16$.

\textbf{Real-world Deployment}
A key challenge in real-world deployment is balancing model capacity and inference latency. Similar to UAV-Flow, we adopt a ground–drone collaborative framework: the VLM and VLA run on a server with an RTX A6000 GPU, while only SLAM~\cite{xu2021fast} and path planning~\cite{zhou2020ego} are executed onboard. The UAV streams 1K first-person video and state information to the ground station via WiFi at 30 Hz, and receives predicted actions in return for closed-loop execution. To mitigate control mismatch caused by communication and inference delay, we adopt a simple \textit{Stop-and-infer} strategy: the UAV waits after executing the previous action until the next action is received.


\begin{table}[t]
\centering
\small
\caption{Performance comparison of different baselines on EP, Air-EQA, OBP\protect\footnotemark, and FLUC\protect\footnotemark tasks.}
\label{tab:main_results}
\setlength{\tabcolsep}{8pt}

\newcommand{\grouprow}[2]{%
\multicolumn{9}{l}{%
    \cellcolor{#1}%
    \parbox[c][2.6ex][c]{0.28\linewidth}{\sffamily\bfseries #2}%
} \\
}
\resizebox{0.9\linewidth}{!}{
\begin{tabular}{lcccccccc}
\toprule
\multirow{2}{*}{Method} & \multicolumn{1}{c}{EP} & \multicolumn{2}{c}{Air-EQA} & \multicolumn{1}{c}{OBP} & \multicolumn{4}{c}{FLUC} \\
\cmidrule(lr){2-2} \cmidrule(lr){3-4} \cmidrule(lr){5-5} \cmidrule(lr){6-9}
 & SR & Acc. & APL & Acc. & SR & OSR & nDTW & NE \\
\midrule

\grouprow{red!20}{VLM + VLA}
GPT-5.4 + OpenVLA             & 18.9 & 71.0 & 37.4 & 72.5 & 13.1 & 28.9 & 12.2 & 7.55  \\
GPT-5.4 + Pi-0.5              & 47.8 & 66.8 & 27.6 & 72.5 & 31.0 & 71.0 & 12.9 & 9.86  \\
Gemini-2.5-Pro + OpenVLA     & 17.1 & 71.4 & 38.6 & 70.4 & - & - & - & -  \\
Gemini-2.5-Pro + Pi-0.5      & 49.4 & 68.1 & 27.4 & 70.4 & - & - & - & -  \\
Qwen3-VL-Max + OpenVLA       & 19.1 & 70.8 & 36.7 & 63.9 & - & - & - & -  \\
Qwen3-VL-Max + Pi-0.5        & 46.7 & 63.9 & 26.5 & 63.9 & - & - & - & -  \\
\midrule

\grouprow{yellow!20}{Human + VLA}
Human + OpenVLA              & 24.8 & 82.3 & 58.3 & 99.1 & - & - & - & - \\
Human + Pi-0.5               & 60.3 & 85.0 & 41.2 & 99.1 & - & - & - & - \\
\midrule

\grouprow{blue!20}{VLM + Human}
GPT-5.4 + Human              & 69.2 & 69.2 & - & 72.5 & - & - & - & -  \\
Gemini-2.5-Pro + Human       & 66.1 & 66.1 & - & 70.4 & - & - & - & -  \\
Qwen3-VL-Max + Human         & 64.3 & 64.3 & - & 63.9 & - & - & - & -  \\
\midrule
\rowcolor{green!20}
\multicolumn{1}{l}{\sffamily\bfseries Human Agent} & 97.3 & 98.2 & - & 99.1 & - & - & - & - \\
\bottomrule
\end{tabular}
}
\vspace{-0.5cm}
\end{table}
\footnotetext[1]{The same VLM yields the same OBP results.}
\footnotetext[2]{The same VLA yields the same FLUC results.}

\section{Experiments}
We conduct a comprehensive evaluation of popular VLMs and VLA models on our benchmark. We adapt VLA models to the fine-grained action space of FLUC and evaluate them in both simulated environments. These methods are also validated on the real-world dataset. We systematically analyze the error sources of UAV agents in active perception tasks to identify the key bottlenecks that limit their performance.

\subsection{Experimental Setup}
\textbf{Baselines.}
We construct different ActiveFly agent baselines by combining different VLM and VLA models. For VLMs, we select three commercial models with strong reasoning ability: GPT-5.4~\cite{singh2025openai}, Gemini-2.5 Pro~\cite{comanici2025gemini}, and Qwen3-VL-Max~\cite{bai2025qwen3}. For VLA models, we adopt the representative discrete-action prediction model OpenVLA~\cite{kim2024openvla} and the continuous-action prediction model Pi-0.5~\cite{intelligence2025pi05}. We modify the output dimension of VLA models to adapt to the UAV action space. We further replace either the VLM or the VLA module with human agents to establish upper-bound performance.

\textbf{Metrics.}
For Air-EQA, we use the accuracy of the multiple-choice question (MCQ) and the accuracy weighted by the length of the path (APL)~\cite{majumdar2024openeqa} to evaluate the accuracy and efficiency of the baseline models. For FLUC, we adopt three widely used metrics from VLA and VLN: Success Rate (SR), Oracle Success Rate (OSR), Navigation Error (NE), and normalized Dynamic Time Warping (nDTW). And we define the success of the language-guided embodied perception (EP) task as
    $ \text{S}_{ep} = \text{S}_{obp} \cdot \text{OS}_{fluc} \cdot \text{S}_{eqa}$,
where $\text{S}_{obp}$, $\text{S}_{eqa}$, $\text{OS}_{fluc} \in \{0, 1\}$ indicate whether the agent correctly answers OBP, Air-EQA, and achieves oracle success in FLUC, respectively. And the SR of EP is the ratio of successful cases.

\begin{figure}[t]
  \centering
  \includegraphics[width=\linewidth]{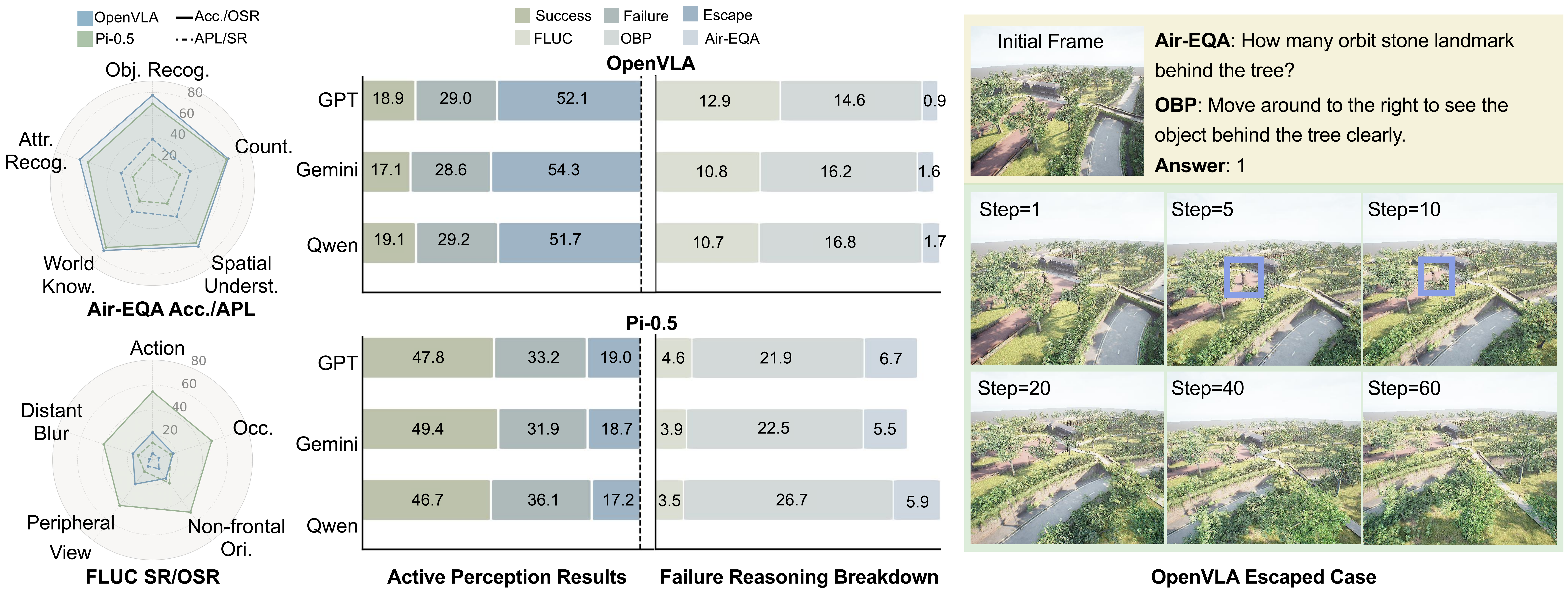}
  \vspace{-0.6cm}
  \caption{\textbf{Category-level performance (left), error breakdown (middle) and case study of \textit{"escaped"} case (right).} The escaped case shows that although the agent deviates from the correct trajectory, it can still acquire the critical visual information (purple box) required to answer the embodied question.}
  \vspace{-0.4cm}
  \label{fig:err_analysis}
\end{figure}

\subsection{Overall Performance}
We report the overall performance on the three tasks in Table~\ref{tab:main_results}. On EP, Pi-0.5-based agents achieve an SR about 30\% higher than OpenVLA-based agents. In contrast, all non-human baselines obtain much higher accuracy on Air-EQA than on EP, mainly due to the \textit{Air-EQA escape} phenomenon (\(\text{S}_{eqa}=1\) but \(\text{S}_{ep}=0\)), which we analyze further in Section~\ref{subsec: err_analysis}. APL is consistently much lower than accuracy for all baselines, especially those built on Pi-0.5, indicating that successful completion often requires long exploration trajectories. On OBP, Qwen3 performs notably worse than GPT-5.4 and Gemini-2.5, suggesting weaker multimodal reasoning and planning ability.

On FLUC, Pi-0.5 achieves 31\% SR and 71.0\% OSR, outperforming OpenVLA by 18\% in SR and by a large margin in OSR. However, both models show poor trajectory similarity, with nDTW around 12\%. OpenVLA achieves lower navigation error, with an NE about 2 meters smaller than Pi-0.5.

Overall, Pi-0.5 is more likely to follow instructions and pass through the target region, which improves oracle success and benefits Air-EQA by exposing the agent to richer scene semantics. However, its longer trajectories reduce efficiency, leading to lower APL and larger navigation errors.

\subsection{Category-Level Performance}
In this section, we present the category-level performance of different VLA models across the three tasks. On Air-EQA, the Pi-0.5-based agent has lower APL than the OpenVLA-based agent across all question categories due to its longer exploration trajectory.
On OBP, the VLM achieves significantly higher planning success rates in distant blur, peripheral view, and elementary trajectory settings than in occlusion and non-frontal orientation scenarios.
On FLUC, Pi-0.5 achieves higher SR and OSR than OpenVLA across all subcategories, while performing comparably to OpenVLA in terms of nDTW. However, the gap between SR and OSR is larger for Pi-0.5, especially on simple tasks such as elementary trajectories. We attribute this to the tendency of Pi-0.5 trajectories to pass through the target position rather than stop precisely at it.

\subsection{Error Analysis}
\label{subsec: err_analysis}
We analyze the error distribution across different VLMs and VLAs. As shown in Figure~\ref{fig:err_analysis}, the dominant error sources are OBP failure and Air-EQA escape. OBP failure accounts for up to 20\% of errors in Pi-0.5-based agents and around 15\% in OpenVLA-based agents.

Air-EQA escape refers to cases where the agent fails to correctly solve OBP or fails to pass through the target position, yet still answers the embodied question correctly. This phenomenon is particularly pronounced in OpenVLA-based agents. 
Through further visualization, we find that this effect arises from the strong scene understanding capability of VLMs: even when the agent trajectory does not pass through the target position, the VLM can still infer the correct answer from the visual semantics accumulated during exploration. 
The relatively low rate of Air-EQA failure further supports this assumption. However, such escape trajectories are often suboptimal and may even be unnecessarily long, as illustrated in Figure~\ref{fig:err_analysis}. In addition, the FLUC failure rate of Pi-0.5 is substantially lower than that of OpenVLA, further confirming the stronger action prediction and language-instruction-following capabilities of Pi-0.5. 

\begin{figure}[t]
  \centering
  \includegraphics[width=\linewidth]{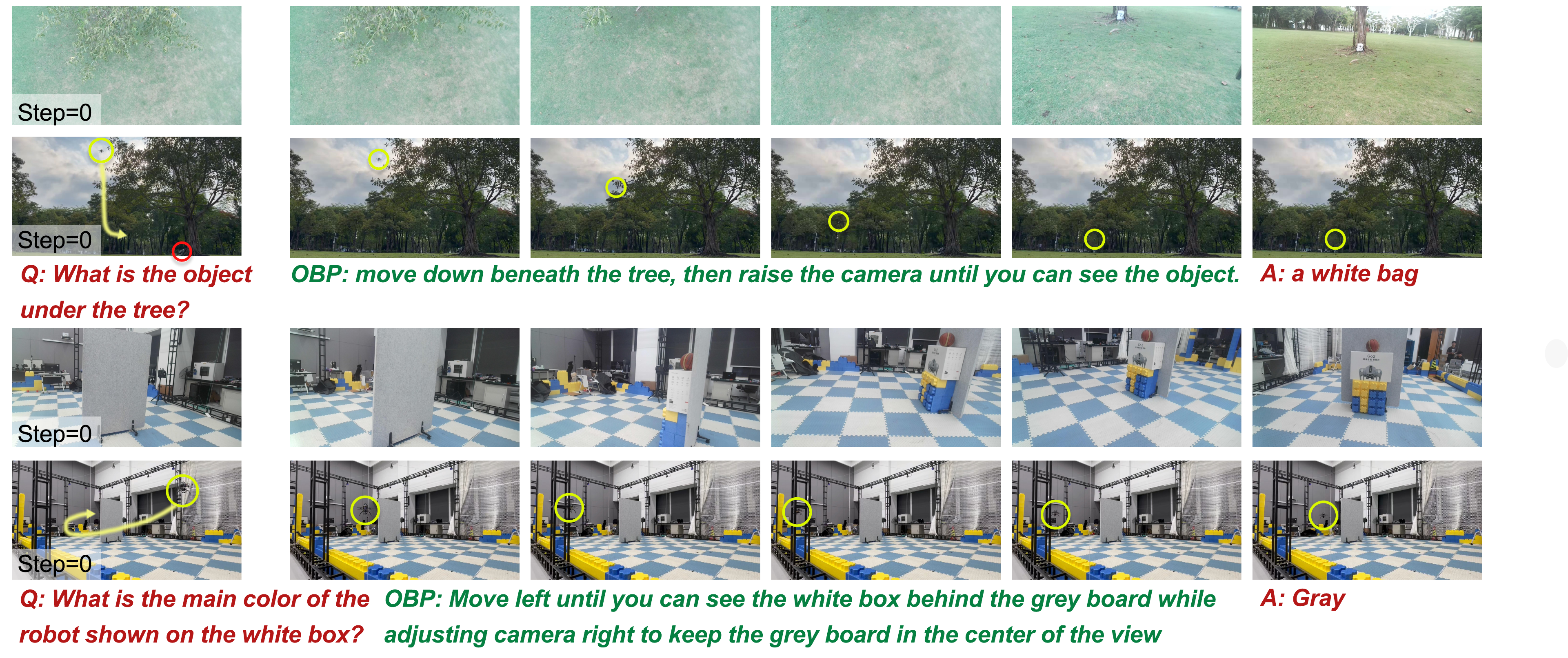}
  \vspace{-0.5cm}
  \caption{\textbf{Visualization of real-world deployment of ActiveFly agent.} The top and bottom rows show the first- and third-person views, respectively.}
  \vspace{-0.5cm}
  \label{fig:real_world_demo}
\end{figure}

\subsection{Real-World Validation}
\label{subsec: real_world_res}
We present the real-world deployment and validation results of the ActiveFly Agent in both indoor and outdoor environments. The agent uses GPT-5.4 for QA and OpenVLA for action prediction. The entire system is implemented in a ground–drone collaborative framework and operates in a closed-loop manner.
As shown in Figure~\ref{fig:real_world_demo}, the UAV agent can autonomously perform reasoning and fine-grained action control, and answer the embodied question.
\begin{wraptable}{r}{0.7\linewidth}   
\vspace{-0.5cm}
\caption{\textbf{System latency breakdown of a single control loop.}}
\label{tab:latency_breakdown}
\renewcommand{\arraystretch}{1.1}
\setlength{\tabcolsep}{6pt}
\resizebox{\linewidth}{!}{
\begin{tabular}{@{}lccccc@{}}
\toprule
 & Image transmission & VLA inference & Motion planning & PID response & Total \\
\midrule
Latency & $\sim120$ &$\sim250$ & $\sim10$ & $\sim10$ & $\sim390$\\
\bottomrule
\end{tabular}
}
\vspace{-0.4cm}
\end{wraptable}
We also report the system latency of a single control loop in Table~\ref{tab:latency_breakdown}. The latency is measured under a 5 Mbps Wi-Fi connection. The primary latency comes from image transmission between the UAV and the ground station, as well as VLA model inference. Nevertheless, the total latency of one control cycle remains within 1 second, indicating that the system can execute embodied perception tasks in a relatively smooth and responsive manner under the common network conditions.

\section{Conclusion}
In this work, we introduce ActiveFly-Bench, the first benchmark bridging high-level EQA and low-level action control for UAV embodied perception. It decomposes embodied perception into three hierarchical tasks and provides 10k FLUC trajectories together with 2.6k QA pairs for reasoning and planning evaluation. We further develop ActiveFly, a closed-loop UAV agent for real-world deployment. Experiments show that current UAV agents still struggle with observation planning, viewpoint adjustment, and robust task completion. We hope ActiveFly-Bench will serve as a useful testbed for future research on UAV embodied perception.


\bibliographystyle{unsrt}
\bibliography{main}

\newpage
\appendix

\section{Technical appendices and supplementary material}

\subsection{Dataset Details}


\subsubsection{Details of Trajectory/Image Curation in Simulator}
\textbf{Advantages of the Selected Simulators:}\\
a) Realistic environmental modeling: Both Beijing and Citypark are built on the Unreal Engine, encompassing diverse architectural styles and street layouts. They feature over a hundred micro-city elements, significantly enriching the semantic information of the acquired embodied agent video data. \\
b) Support for aerial agents: Both simulators are equipped with the built-in AirSim plugin, facilitating the control of aerial agents. \\
c) Existing route references: Previous research on vision-language navigation has been conducted in these simulators \cite{zhu2020vision}, enabling us to obtain specific route coordinates and instruction data. Although most of these routes cannot be directly utilized (e.g., the majority of flight paths in AerialVLN contain numerous meaningless, repetitive maneuvers and lack logical progression, conflicting with the purpose of this benchmark.), they provide valuable reference for our data collection process.\\
\textbf{Drone Settings in the Simulator}: The drone is equipped with a gimbal-mounted camera supporting pitch movements ranging from 0 to 90 degrees.\\
\textbf{Collection Standards}: The drone agents were operated by a total of five experienced pilots to ensure the rationality of the flight maneuvers. We carefully balanced the efficiency and diversity of the flight paths to ensure that the collected data comprehensively represents drone motion patterns in real-world Vision-Language-Action (VLA) tasks. This approach serves to validate the VLA and embodied Question \& Answer (EQA) capabilities of Video Large Language Models (Video-LLMs).

\subsubsection{Details of Trajectory/Image Curation in Real World}

For the real-world tasks, our task categories are identical to those used in simulation. For each task, we collected a total of 4,000 indoor and outdoor expert action trajectories, and the trajectory design also follows the simulation setup: it consists of eight atomic actions—move left, move right, turn left, turn right, move up, move down, camera up, camera down—and five VLA actions: occlusion, central\_view, fuzzy, orientation, normal. Since a real drone cannot achieve the precise and efficient motion of a simulated drone, manually controlling the drone to collect a large number of images is impractical. Therefore, we designed an automated data collection pipeline based on expert demonstrations, detailed as follows:\\

First, at the expert trajectory collection level, for each task our professional pilot flies an expert trajectory according to the task requirements. We extract the odometry data (x, y, z, yaw) and gimbal state data (gimbal-state, including roll, pitch, yaw) from the recorded ROS bag. To distill the core information that characterizes the motion trend of the trajectory from the dense raw trajectory points, we perform adaptive sampling on the trajectory in specified planes (e.g., X‑Y, X‑Z, Y‑Z): we first compute the rate of curvature change at each point to identify key motion phases such as turning and speed changes, while simultaneously imposing a minimum physical distance constraint between adjacent points to avoid overly dense sampling. By adjusting the curvature threshold and the minimum spacing, the sampling granularity can be flexibly controlled (defaulting to 8 key points). The resulting key-points are represented as a five-tuple (x, y, z, yaw, gimbal-state), capable of effectively capturing the spatial path and attitude variation trend with only a small number of points.
\\

Next, to increase dataset diversity, we perturb these key points: Gaussian noise with zero mean and a standard deviation of 0.1 m is added to the position coordinates, and Gaussian noise with zero mean and a standard deviation of 0.05 rad is added to the yaw angle. This yields multiple new trajectories that slightly differ in position and orientation yet share the overall motion pattern of the expert trajectory. This perturbation strategy enriches the state-action distribution for subsequent policy learning without destroying task semantics. Finally, at the control level, we realize trajectory tracking along two dimensions. For the drone body, we modify EGO-Planner so that it can constrain height (z) and yaw angle, and use a PID controller to drive the drone to move smoothly between adjacent key-points. For the gimbal, we design a dedicated servo controller according to its motion characteristics, making it follow the gimbal state corresponding to the key points in a coordinated manner. The two controllers work together so that the drone moves along the expert trajectory while the gimbal maintains a camera angle consistent with the task requirements, thereby automatically completing the image data collection.\\

Through this pipeline, we can use a small number of human demonstrations as seeds to generate annotated training samples at scale, providing a data foundation for subsequent learning.\\



\subsubsection{Instruction Details}
Consistent with the dataset structure presented in the main text, the text instructions in our dataset are categorized into Atomic Actions and VLA Tasks.
For Atomic Actions, the main body of the instruction comprises one of ten single actions: "move up", "move down", "move forward", "move backward", "move left", "move right", "camera up", "camera down", "camera left", "camera right".
Meanwhile, the instructions for VLA Tasks are composed of atomic actions as the minimal units.
This compositional approach ensures that the fine-tuning and testing phases of VLA Tasks can maximally benefit from those of the Atomic Action phase.
Specific examples are provided in Figure \ref{fig:more_examples_obp}.\\

All instructions were manually annotated by the pilots involved in the trajectory/image curation process to guarantee precise alignment between the text instructions and their corresponding trajectories.

\subsection{Platform Details}

\subsubsection{Hardware Configuration}
The UAV hardware platform is constructed to support agile flight and high-quality data collection in complex environments. The core airframe utilizes a compact 280-mm wheelbase carbon fiber design. Propulsion is driven by T-Motor F90 motors, providing sufficient thrust for responsive maneuvers. Onboard computation is handled by an Intel NUC 13 ANKB mini PC, which processes real-time perception, trajectory planning, and data logging. Low-level flight control is executed by a flight control unit running the PX4 Autopilot. 

For environmental perception, a Livox Mid-360 3D LiDAR is integrated to provide omnidirectional depth measurements. To capture stable and viewpoint-adjustable visual observations, a high-resolution camera is mounted on an XF C-200T 3-axis gimbal. This gimbal system allows independent adjustment of the camera's pitch, roll, and yaw. Such configuration directly supports the collection of diverse observation behaviors and captures the precise gimbal states required for the real-world trajectory curation detailed in Appendix A.3.3.

\subsubsection{Navigation and Control System}
The autonomous navigation system integrates efficient state estimation and local planning modules to ensure safe obstacle avoidance during flight. State estimation is powered by FAST-LIO2, a tightly-coupled LiDAR-inertial odometry framework that fuses LiDAR scans and IMU measurements. It provides high-frequency, low-latency pose estimates, maintaining accurate localization during continuous and agile flight. 

Real-time obstacle avoidance and trajectory generation are achieved using Ego Planner. By operating directly on LiDAR point clouds without constructing a global Euclidean Signed Distance Field (ESDF), Ego Planner generates smooth and collision-free local paths with minimal computational overhead. The integration of FAST-LIO2 and Ego Planner guarantees that the UAV can autonomously navigate through cluttered spaces and safely execute the required flight maneuvers during the automated data collection process.

\subsection{EQA Details}


\subsubsection{Details of EQA Generation}
\label{appendix: Details of EQA Generation}

As described in the main text, all Embodied Question \& Answer (EQA) instances are categorized into five types: "Object Recognition", "Attribute Recognition", "World Knowledge", "Localization", and "Spatial Understanding".
The EQA tasks are formulated as Multiple Choice Questions (MCQs), consisting of one correct option and three distractor options;
the model is required to select the correct answer from these four choices.
Each EQA is assigned a unique "question\_id" and corresponds to a unique test sample based on the "data\_id".
Notably, not every test sample is associated with an EQA, nor does every sample encompass all types of EQAs.
Only scenarios and questions that strictly meet our predefined criteria are adopted.

\textbf{EQA Criteria:}\\
a) Blind Screening Pass: This benchmark focuses on evaluating the model's embodied capability to capture specific visual information through actions.
Therefore, all EQAs should be questions that the model can only answer after demonstrating its embodied capabilities.
Specifically, an EQA must not be correctly answerable from the initial observation position without any movement.\\
b) Unambiguity Principle: Each EQA possesses only one exclusively correct answer.
Since Vision-Language Models (VLMs) generate answers based on historical observations, the three misleading options in the EQA must not introduce any potential confusion with other objects observable along the entire trajectory.\\
c) Indisputability Principle: The correctness and validity of all EQAs are strictly based on human judgment.
Therefore, all personnel involved in the creation and evaluation of the EQAs must reach a unanimous consensus on each question and its corresponding correct answer.\\

In the initial phase of formulating multiple EQAs for each data sample, state-of-the-art commercial VLMs were evaluated for automated EQA generation.
However, the majority of the VLM-generated EQAs required major human refinement.
After carefully weighing efficiency and cost, all EQAs were ultimately authored and evaluated manually.
During the attempts at VLM-based EQA generation, we observed that Large Language Models (LLMs) still struggle with comprehending complex urban environments within videos.
These limitations become particularly pronounced when attempting to interpret the motion patterns of first-person agents and when applying the aforementioned EQA criteria to video understanding.

\subsubsection{Analysis of Failed EQAs Judged by a Single VLM and Corresponding Solutions}
\label{Analysis of Fail EQA judged by a single VLM and Corresponding Solution}

\begin{figure}[htbp]
    \centering
    \includegraphics[width=\textwidth]{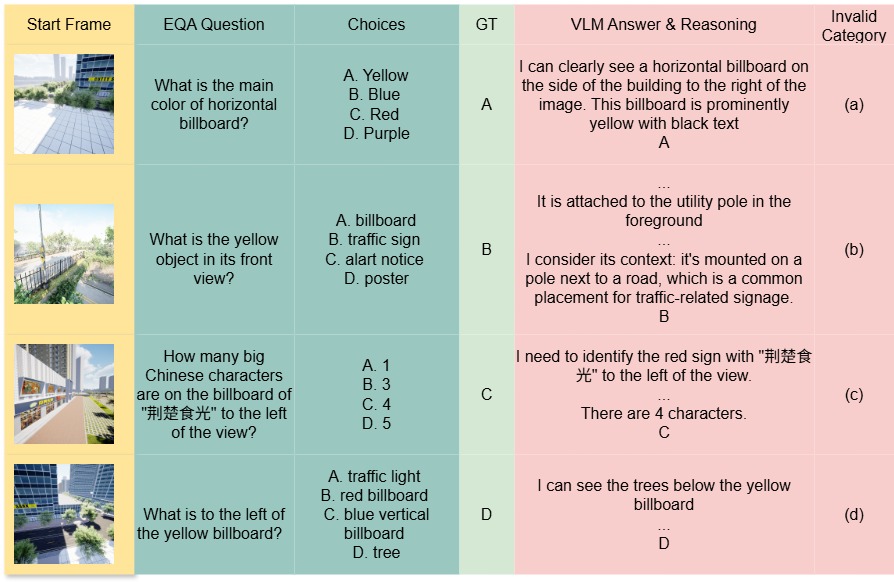}
    \caption{Example of Invalid EQA}
    \label{fig:Example of Invalid EQA}
\end{figure}

After completing the first manual iteration of EQAs for all simulated data, we conducted a blind screening using Gemini 3.1 Pro: we formulated a scenario where the model acts as a static observer.
The EQA and the first frame of the corresponding data were fed into the model, prompting it to return a single option in a standardized format.
If the model answered correctly, the EQA was flagged as invalid.
As a result, 33\% of the EQAs were filtered out.
We analyzed the failure modes of these invalid EQAs, categorizing and summarizing them as follows.
These failures stem from both inherent quality issues in the manually crafted EQAs and the features of the VLMs themselves.
Examples of each failure mode are illustrated in Figure \ref{fig:Example of Invalid EQA}.\\

\textbf{a) Object Appears/Partially Appears in the Start Frame:} When the target object is fully or partially visible in the start frame, the VLM might correctly answer the question even under extremely poor viewing angles (e.g., severe occlusion or oblique angles).
These are considered incidental cases and require human judgment to determine whether they should be retained or revised.
Furthermore, even if the object is only partially visible, certain questions (e.g., querying the object's color) can still be answered directly.
Such invalid EQAs must be discarded.\\

\textbf{b) Guessing Answers Based on Scene and Commonsense:} Even if the start frame lacks a valid observation of the target object, the VLM might guess the most plausible option based on semantic information extracted from the scene.
Addressing this issue requires modifying the distractor options into choices that cannot be easily eliminated using mere scene context.
However, further evaluation revealed that modifying distractors only mitigates this issue.
Regardless of the option design, VLMs actively utilize scene context to eliminate unlikely choices and make guesses, occasionally answering EQAs correctly by chance.
If human evaluators determine that such an EQA still strictly requires "embodied actions to supplement necessary information," it is retained as a valid EQA.\\

\textbf{c) Information Leakage:} Imprecise or non-compliant EQAs might inadvertently suffer from information leakage.
For instance, the question text of the third example in Figure \ref{fig:Example of Invalid EQA} has already exposed the number of letters on the target object.
Such EQAs must be either heavily refined or entirely removed.\\

\textbf{d) Illogical Reasoning by the Model:} Models occasionally arrive at the correct answer through illogical means.
For example, a model might rely on pure random guessing, or, if the target object (such as a specific tree) is absent in the start frame, it might anchor its answer on other similar objects (e.g., another tree) present in the scene (Figure \ref{fig:Example of Invalid EQA}).
These cases require careful manual identification and should be retained.

As demonstrated above, EQAs cannot be filtered through simple binary judgments: some necessitate deletion, others require refinement, and those still meeting the EQA criteria should be retained.
Consequently, a more comprehensive VLM-based Blind Screening pipeline was deployed to address the aforementioned failure modes and to ensure the validity of the EQAs in this benchmark, as detailed in Section \ref{VLM-based Blind Screening of EQA}.

\subsubsection{VLM-based Blind Screening of EQA}
\label{VLM-based Blind Screening of EQA}
The complete VLM-based Blind Screening pipeline utilized in this benchmark is outlined below:

\begin{itemize}
    \item \textbf{Multi-Model Voting:} The single-model judgment was replaced with a multi-model voting mechanism.
Only when all models answer correctly on the start frame is the EQA flagged as ``invalid''.
This effectively minimizes the probability of models guessing the correct answer by chance.
The models employed include Gemini\_2.5\_Flash, GPT\_5.4\_Nano, and Qwen3-VL\_Flash.
    
    \item \textbf{``IDK'' and ``Action Needed'' Labels:} An ``I don't know'' (IDK) option was introduced into the prompt, empowering the model to assess whether the question is answerable.
Furthermore, the concept of ``Embodied'' was explicitly defined in the prompt, and an ``Action needed'' option was added to allow the model to gauge its self-confidence in answering.
    \item \textbf{Chain of Thought (CoT) Output:} The models are required to output their Chain of Thought (CoT), thereby ensuring the interpretability of their answers.
    \item \textbf{Human Cross-Refinement of EQAs:} Based on the CoT generated in the previous step, annotators exchange their EQAs and conduct rigorous peer reviews.
    \item \textbf{Re-evaluation of Refined EQAs:} The CoT from this stage is utilized to determine whether to discard the previously flagged invalid EQAs.
This process allows for the retention of certain edge cases that failed the EQA evaluation but still fundamentally align with the EQA criteria.
\end{itemize}

To ensure that the CoT reasoning process does not interfere with the judgment of answer correctness, the script utilizes an \texttt{extract\_answer()} function to isolate the model's output options, demarcated by the \texttt{<answer>} tags.
Algorithm \ref{alg:eqa_filter} demonstrates the partial core code and prompt of this script.

\begin{algorithm}[tb]
\caption{EQA Filter via Multi-Model Voting}
\label{alg:eqa_filter}
\begin{algorithmic}[1] 
\REQUIRE EQA dataset $\mathcal{D}$, Set of VLMs $\mathcal{M} = \{M_{\text{Gemini}}, M_{\text{GPT}}, M_{\text{Qwen}}\}$, Max retries $N_{\text{retry}}$
\ENSURE Set of invalid EQA IDs $\mathcal{I}$, Evaluation logs $\mathcal{L}$
\STATE Initialize $\mathcal{I} \leftarrow \emptyset$, $\mathcal{L} \leftarrow \emptyset$

\STATE \textbf{Define Prompt Template} $\mathcal{P}(q)$:
\STATE \hspace{0.5cm} \parbox{0.88\linewidth}{%
    \vspace{0.15cm} \small \textit{%
    You are an expert visual QA assistant.
    Look closely at the provided image and answer the question.
    Your task is to determine the correct choice based on the given image.
    \\
    If the image does not contain enough clear information to definitively answer the question, you MUST choose ``I don't know''.
    \\
    Alternatively, if you can guess a vague answer but your confidence is low, and you determine that taking an action to change the camera's position or pose would help acquire more information to answer accurately, you MUST choose ``action needed''.
    \\
    If you can confidently determine the correct answer based on this image, please provide your final choice among ``A'', ``B'', ``C'', ``D''.
    \\
    First, think step-by-step and provide your reasoning (Chain of Thought).
    \\
    Then, you MUST output your final choice wrapped between \texttt{<answer>[Your Choice]</answer>} tags.
    Do not include any other punctuation, explanations, or extra text.
    (This rule must not be violated, as we will use regex to extract the answer from these tags.) \\
    The final choice MUST be exactly one of: ``A'', ``B'', ``C'', ``D'', ``I don't know'', or ``action needed''.
    \\
    restricted format: \\
    Reasoning: [your step-by-step analysis here] \\
    \texttt{<answer>A</answer>} \\
    Question: \{q\}
    } \vspace{0.15cm}
}

\FOR{each EQA instance $x \in \mathcal{D}$}
    \STATE Extract initial frame $V_0$, question $q$, and ground truth $a_{gt}$ from $x$
    \STATE $C \leftarrow \emptyset$ \COMMENT{Track models that answered correctly}
    
    \FOR{each model $M \in \mathcal{M}$ in parallel}
        \STATE $ans_M \leftarrow \text{None}$
        \FOR{$i = 1$ \TO $N_{\text{retry}}$}
            \STATE $r_M \leftarrow \text{GenerateContent}(M, V_0, \mathcal{P}(q))$
            \STATE $ans_M \leftarrow \text{RegexExtract}(r_M, \texttt{<answer>(.*?)</answer>})$
            \IF{$ans_M \neq \text{Error}$}
                \STATE \textbf{break}
            \ENDIF
        \ENDFOR
        \STATE Append CoT and $ans_M$ to $\mathcal{L}$
  
        \IF{$ans_M == a_{gt}$}
            \STATE $C \leftarrow C \cup \{M\}$ \COMMENT{Model guessed correctly on initial frame}
        \ENDIF
    \ENDFOR
    
    \IF{$|C| == |\mathcal{M}|$}
        \STATE $\mathcal{I} \leftarrow \mathcal{I} \cup \{x.\text{id}\}$ \COMMENT{Invalid if all models answer correctly}
    \ENDIF
\ENDFOR
\RETURN $\mathcal{I}, \mathcal{L}$
\end{algorithmic}
\end{algorithm}

\subsubsection{More EQA Examples}
\begin{figure}[htbp]
    \centering
    \includegraphics[width=\textwidth]{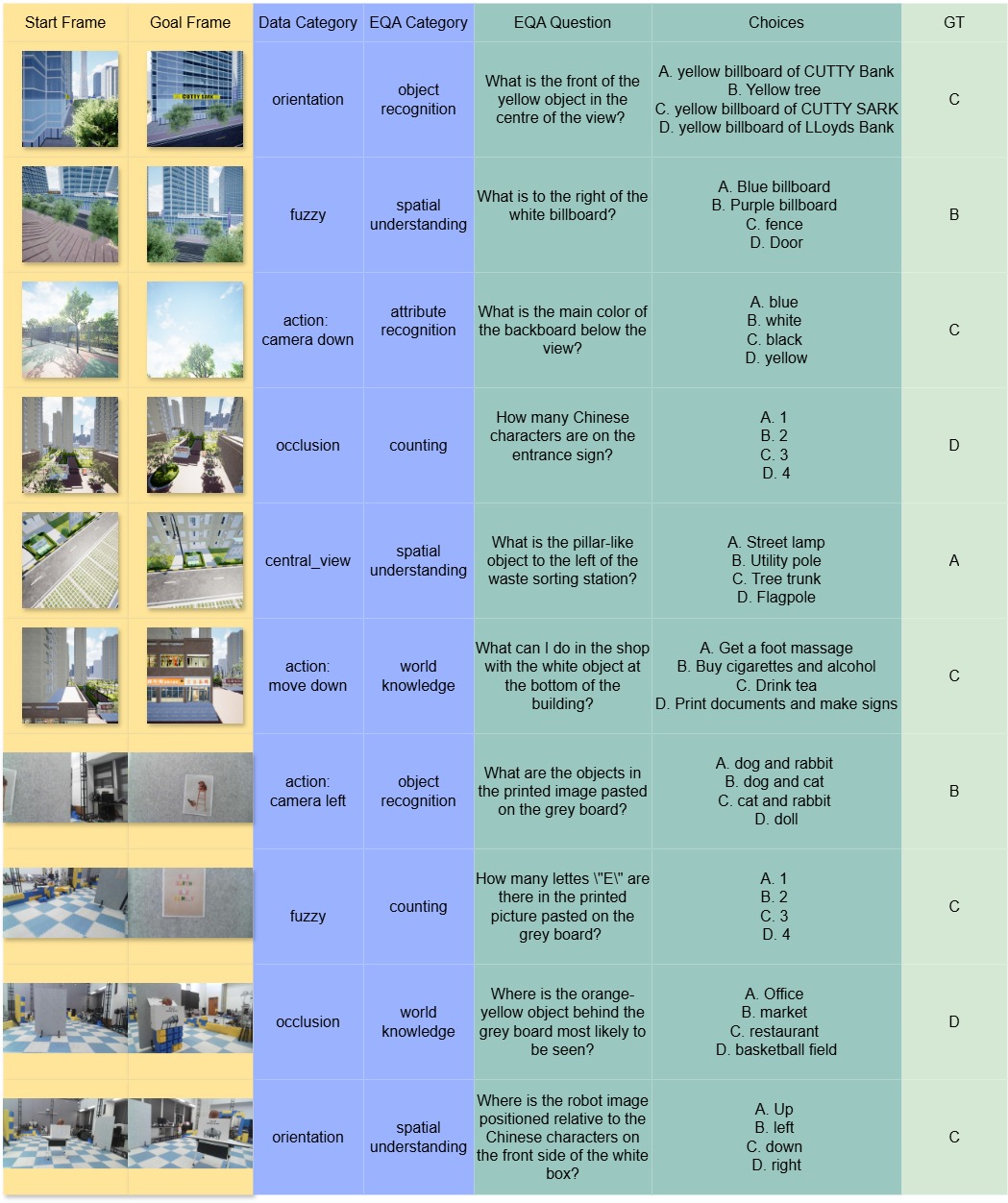}
    \caption{More Examples of EQA}
    \label{fig:more_examples_eqa}
\end{figure}

To better illustrate the proposed EQAs, we provide additional examples encompassing all task and question types across both simulators and the real world.
These are showcased in Figure \ref{fig:more_examples_eqa}.

\subsection{OBP Details}
Observation Behavior Planning (OBP) is presented in a QA format.
The standard template for the question is \textit{"What action is needed to answer the question: \"<EQA question>\"}. The choices consist of the actual trajectory instruction as the correct answer, alongside three misleading instructions that are either physically impossible to execute or would result in a trajectory that fails to answer the EQA question. Figure \ref{fig:more_examples_obp} presents illustrative examples of OBP.

\subsubsection{Details of OBP Generation}
OBP generation must similarly adhere to the unambiguity and indisputability principles applied in EQA generation (Appendix \ref{appendix: Details of EQA Generation}). Furthermore, preventing information leakage is critical in OBP generation. In many data samples, the \textit{object name} and the \textit{EQA question} intrinsically contain spatial cues or hints that inadvertently reveal the answer, as depicted by the third example in Figure \ref{fig:more_examples_obp}.: The name of target object has already indicated the answer. When necessary, the text of the OBP question and choices must be meticulously refined to ensure the problem's validity.
Crucially, an effective OBP relies simultaneously on the essential information present in both the start frame image and the text-based target description.
If the correct action instruction can be deduced from either the text or the image alone, the OBP fails to genuinely evaluate the model's embodied capabilities.

\subsubsection{More OBP Examples}
\begin{figure}[htbp]
    \centering
    \includegraphics[width=\textwidth]{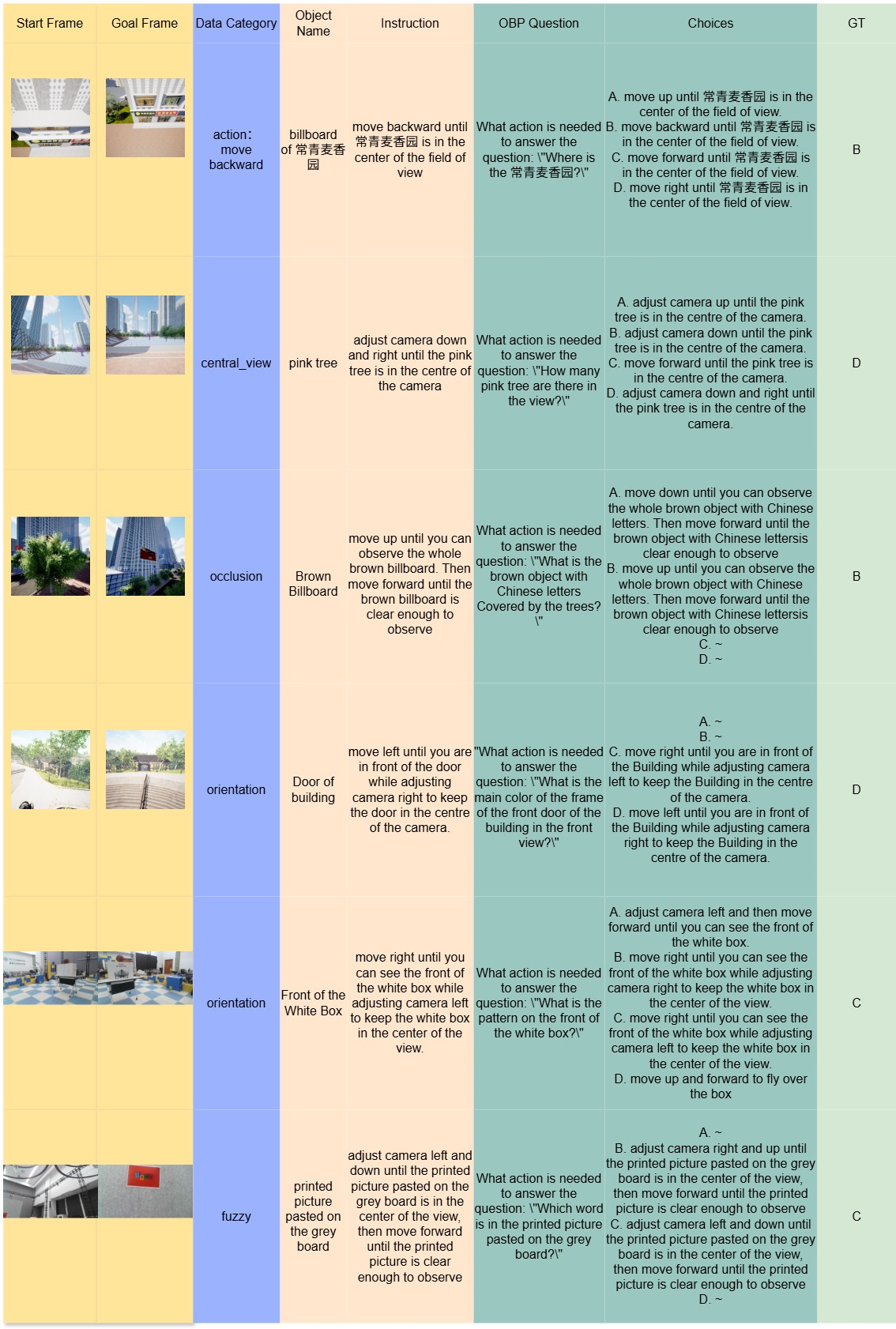}
    \caption{More Examples of OBP}
    \label{fig:more_examples_obp}
\end{figure}

To further clarify the proposed OBP, we provide additional examples encompassing all task types across both simulators and the real world.
These are presented in Figure \ref{fig:more_examples_obp}.

\subsection{Experiment Details}
\subsubsection{Metrics}
Let a predicted trajectory be denoted as 
\begin{equation}
    \hat{\tau} = \{\hat{\texttt{x}}_1, \hat{\texttt{x}}_2, \dots, \hat{\texttt{x}}_T\},
\end{equation}
where $\hat{\texttt{x}}_t$ is the UAV pose at step $t$, and let the ground-truth trajectory be
\begin{equation}
    \tau^* = \{\texttt{x}^*_1, \texttt{x}^*_2, \dots, \texttt{x}^*_N\}.
\end{equation}
Here, each pose $\texttt{x}=[\texttt{x}_\mathrm{loc}:\texttt{x}_\mathrm{ori}]$ includes both position $\texttt{x}_\mathrm{loc}$ and orientation $\texttt{x}_\mathrm{ori}$. We use $d_\mathrm{Man}(\cdot,\cdot)$ and $d_\mathrm{Euc}(\cdot,\cdot)$ to denote the Manhattan and Euclidean distance, respectively.

\textbf{The success rate (SR)} follows the definition used in aerial VLN~\cite{liu2023aerialvln}. However, since FLUC targets short-horizon, viewpoint-aware navigation, we additionally take the UAV's yaw and pitch angles into account when determining whether the target has been successfully reached. Accordingly, SR is defined as follows:
\begin{equation}
\mathrm{SR} =
\frac{1}{M}
\sum_{i=1}^{M}
\mathbf{1}\!\left[
d_{\mathrm{Euc}}\!\left(\hat{\texttt{x}}^{(i)}_{T,\mathrm{loc}}, \texttt{x}^{*(i)}_{N,\mathrm{loc}}\right) < \delta_{\mathrm{loc}}
\ \land\
d_{\mathrm{Man}}\!\left(\hat{\texttt{x}}^{(i)}_{T,\mathrm{ori}}, \texttt{x}^{*(i)}_{N,\mathrm{ori}}\right) < \delta_{\mathrm{ori}}
\right],
\end{equation}
where $M$ is the number of evaluation samples. We set $\delta_{\mathrm{loc}}=3$ and $\delta_{\mathrm{ori}}=10^\circ$.

\textbf{Oracle Success Rate (OSR)} measures whether the predicted trajectory reaches the target pose at any step. Under the viewpoint-aware FLUC setting, a trajectory is regarded as oracle-successful if there exists at least one step whose position and orientation are both sufficiently close to the ground-truth target pose. Formally, OSR is defined as
\begin{equation}
\mathrm{OSR}
=
\frac{1}{M}
\sum_{i=1}^{M}
\mathbf{1}\!\left[
\exists\, t \in \{1,\dots,T\},
\ d_{\mathrm{Euc}}\!\left(\hat{\texttt{x}}^{(i)}_{t,\mathrm{loc}}, \texttt{x}^{*(i)}_{N,\mathrm{loc}}\right) < \delta_{\mathrm{loc}}
\ \land\
d_{\mathrm{Man}}\!\left(\hat{\texttt{x}}^{(i)}_{t,\mathrm{ori}}, \texttt{x}^{*(i)}_{N,\mathrm{ori}}\right) < \delta_{\mathrm{ori}}
\right].
\end{equation}
\textbf{Navigation Error (NE)} is defined as the distance between the final predicted pose and the ground-truth final pose:
\begin{equation}
    \mathrm{NE} = \frac{1}{M}\sum_{i=1}^{M} d_\mathrm{Euc}\left(\hat{\texttt{x}}^{(i)}_{T,\mathrm{loc}}, \texttt{x}^{*(i)}_{N,\mathrm{loc}}\right).
\end{equation}

\textbf{Normalized Dynamic Time Warping (nDTW)} evaluates trajectory similarity by comparing the predicted trajectory with the ground-truth trajectory using Dynamic Time Warping (DTW). Let $\mathrm{DTW}(\hat{\tau}, \tau^*)$ denote the DTW distance between the two trajectories. Then nDTW is defined as
\begin{equation}
    \mathrm{nDTW} = \frac{1}{M}\sum_{i=1}^{M} \exp\!\left(-\frac{\mathrm{DTW}(\hat{\tau}^{(i)}, \tau^{*(i)})}{\eta \, |\tau^{*(i)}|}\right),
\end{equation}
where $|\tau^{*(i)}|$ is the length of the ground-truth trajectory, and $\eta$ is the normalizing success threshold. We set $\eta=1$.

\textbf{Accuracy weighted by Path Length (APL)} jointly evaluate correctness and efficiency.APL rewards correct answers obtained with shorter exploration trajectories.
Let $a_i \in \{0,1\}$ denote whether the answer to the $i$-th question is correct, and let $L_i$ denote the length of the corresponding exploration trajectory. Then APL is defined as
\begin{equation}
\mathrm{APL}
=
\frac{1}{M}
\sum_{i=1}^{M}
a_i \cdot
\frac{|\tau^{*(i)}|}{\max\left(|\hat{\tau}^{(i)}|, |\tau^{*(i)}|\right)},
\end{equation}
In practice, a higher APL indicates that the agent can answer the question correctly with a shorter exploration path, and therefore reflects better overall efficiency in embodied perception.


\newpage

\newpage

\end{document}